\documentclass[11pt,letterpaper]{article}
\usepackage[utf8]{inputenc}
\usepackage[T1]{fontenc}
\usepackage{amsmath}
\usepackage{mathptmx}
\usepackage{graphicx}
\usepackage{array}
\usepackage[letterpaper,margin=1in]{geometry}
\usepackage{newunicodechar}
\usepackage[hidelinks]{hyperref}
\usepackage{xurl}
\usepackage{setspace}
\graphicspath{{figures/}}
\newunicodechar{π}{\ensuremath{\pi}}\newunicodechar{φ}{\ensuremath{\varphi}}\newunicodechar{θ}{\ensuremath{\theta}}\newunicodechar{λ}{\ensuremath{\lambda}}
\newunicodechar{×}{\ensuremath{\times}}\newunicodechar{−}{\ensuremath{-}}\newunicodechar{±}{\ensuremath{\pm}}
\newunicodechar{∥}{\ensuremath{\parallel}}\newunicodechar{≈}{\ensuremath{\approx}}\newunicodechar{≥}{\ensuremath{\geq}}
\newunicodechar{≤}{\ensuremath{\leq}}\newunicodechar{°}{\ensuremath{^\circ}}\newunicodechar{⊤}{\ensuremath{\top}}
\newunicodechar{⟨}{\ensuremath{\langle}}\newunicodechar{⟩}{\ensuremath{\rangle}}\newunicodechar{·}{\ensuremath{\cdot}}
\newunicodechar{∑}{\ensuremath{\sum}}\newunicodechar{√}{\ensuremath{\surd}}\newunicodechar{∘}{\ensuremath{\circ}}
\begin{document}
\begin{center}
{\Large\bfseries Vision-Guided Reorientation and Color Sorting of Writing Utensils Using a Low-Cost 4-DoF Arm}\\[5pt]
Anirudh Rangarajan\textsuperscript{1*}, Bibit Bianchini\textsuperscript{2}\\[5pt]
\emph{1. Dougherty Valley High School, 10550 Albion Road, San Ramon, California 94582, United States of America}\\[5pt]
\emph{2. University of Pennsylvania, 3451 Walnut Street, Philadelphia, Pennsylvania, 19104, United States of America}\\[5pt]
\emph{* Corresponding author email: \href{mailto:anirudh.rangarv@gmail.com}{{anirudh.rangarv@gmail.com}}}
\end{center}

\textbf{Published version.} This article is published in the American Journal of Student Research: Rangarajan A, Bianchini B. Vision-Guided Reorientation and Color Sorting of Writing Utensils Using a Low-Cost 4-DoF Arm. Am J Stud Res. 2026;4(6):74-88. \url{https://doi.org/10.70251/HYJR2348.467488}. The published version is open access under a Creative Commons Attribution 4.0 International (CC BY 4.0) license. This arXiv version is the accepted manuscript in the authors' own formatting; an earlier version of this work was posted as arXiv:2608.15968v1 under the title ``Tabletop Pen Manipulation With a Vision-Guided 4-DoF Arm.''

\textbf{ABSTRACT}

Low-cost four-degree-of-freedom (DoF) arms are among the most accessible robotic platforms. But they are one joint short for picking up in situations where objects are at arbitrary orientations, a task that appears to require five degrees of freedom: the planar position (x and y), the height (z), a wrist rotation to align the gripper with the object, and gripper actuation, of which a four-DoF arm lacks the wrist rotation. This work uses perception and motion planning so that such an arm, a roughly \$200 Waveshare RoArm-M2-S under a fixed overhead camera, can detect and color-sort writing utensils without that joint. A YOLO11n-OBB (You Only Look Once, oriented bounding box) detector locates each writing utensil; camera intrinsics and an ArUco reference pose convert its pixel coordinates to robot coordinates; and a color classifier labels it. The detected orientation angle determines the motion strategy: utensils close to the arm's fixed approach direction are picked up directly, and those at steeper angles are reoriented via corrective sweeps until they are graspable, after which they are picked up and sorted into the assigned color bin. In 196 single-utensil trials, the 98 that started beyond 45 degrees from that direction all became graspable within four sweeps (130 total). Overall, 176 (89.8\%) ended in the correct bin; 19 of 20 failures were slipped or missed grasps, one a mislabeled color. For the seven utensils tested in this setup, corrective sweeps brought every steeply misaligned utensil into a graspable orientation on an arm with no wrist joint.

\hypertarget{keywords}{%
\section{\texorpdfstring{\textbf{KEYWORDS}}{KEYWORDS}}\label{keywords}}

Robotics; manipulation; computer vision; sorting; intrinsics;
extrinsics; 4-DoF

\hypertarget{introduction}{%
\section{\texorpdfstring{\textbf{INTRODUCTION}}{INTRODUCTION}}\label{introduction}}

Low-cost robotic arms with four degrees of freedom are among the most accessible manipulation platforms available to students, small labs, and independent researchers. They are affordable, compact, and well-suited to a broad class of structured tasks; the arm used in this work, the Waveshare RoArm-M2-S [1], retails for roughly \$200. However, each additional degree of freedom in a robot arm expands its range of motion, and those missing degrees become a practical constraint when objects lie at arbitrary orientations.

Pen color-sorting is a representative case. Pens lie at arbitrary angles and must be identified by color, so picking one up requires the gripper to match its orientation, sometimes by rotating the wrist. Without that DoF, a 4-DoF arm needs another path. In this work, the arm compensates through perception and planning. A YOLO11n-OBB (You Only Look Once, oriented bounding box) detector provides each pen's orientation angle directly, and a color classifier identifies its label. The motion planner then uses the orientation to either attempt a direct perpendicular grasp or to perform one or more series of nudges, using short lateral sweeps of the pen to push it into a graspable orientation.

This work investigates how well vision-guided perception and geometry-driven motion planning can enable a 4-DoF arm to accomplish a color pen-sorting task that routes each detected utensil into one of four bins (blue, red, green, or grayscale), a task that nominally demands five degrees of freedom. The RoArm-M2-S has five serial-bus servos driving four actuated joints: base rotation, shoulder pitch (two servos on one axis), elbow pitch, and the gripper [1, 2]. The first three set the gripper's position (x, y, z); the fourth only opens and closes the jaws. With no wrist joint, the jaws always close across the arm's radial direction, so a pen must lie roughly along that line to be gripped across its width; aligning the jaws with a pen at an arbitrary angle would need one more actuated axis, a rotation about the vertical. The task therefore calls for four pose degrees of freedom plus gripper actuation, and this arm supplies three plus the gripper [3]. The arm is not underactuated, since every joint it has is driven; it is missing a wrist axis.

Pushing is a basic manipulation primitive alongside grasping, and using it to set an object up for a grasp has a long history. Early work on robotic pushing treated the mechanics of a flat object pushed across a table. Mason [4] showed that an object pushed at a single point either rotates or slides, depending on where the line of the push passes relative to the object's center of friction, which is what makes a push predictable enough to plan. Push-to-grasp methods then used pushing to make a grasp possible, by planning a rotation of the object before the grasp [5], by sweeping the object into the hand with push-grasps that absorb pose uncertainty [6], by learning when to push and when to grasp in clutter [7], and by sliding thin objects to a table edge where a compliant hand can take them [8]. In every one of those systems, the arm could also turn its hand, so pushing was one option among several, and a planner or a learned policy chose the push. On the sensing side, many camera-guided arms use visual servoing, correcting from the image while the arm moves [9], and the same approach has been used on a low-DoF arm, where a 4-DoF palletizing arm sorted objects without a calibrated camera [10].

As far as we know, this combination has not been reported before. The arm has no wrist rotation at all, so pushing is the only way to line up a grasp, and each motion is driven by one detection rather than by a feedback loop or a learned policy. The angle from an oriented-box detector decides between a direct grasp and a sweep-then-grasp sequence, the direction and length of the push follow from one geometric rule, and a calibrated pixel-to-robot mapping says where to push. Nothing is searched, nothing is learned beyond the detector, and nothing is corrected while a motion runs. That pipeline and its evaluation over 196 logged trials, in which 176 (89.8\%) ended in the bin for the pen's color, are the contribution of this paper.

\hypertarget{methods-and-materials}{%
\section{\texorpdfstring{\textbf{METHODS AND
MATERIALS}}{METHODS AND MATERIALS}}\label{methods-and-materials}}

Code Reference. All source code and the repository files are publicly
available at:\\
\href{https://github.com/Anirudhpro/4DoF_vision_robotic_pen_sorting}{github.com/Anirudhpro/4DoF\_vision\_robotic\_pen\_sorting}
{[}11{]}.

The system pipeline has three stages: oriented bounding box detection, pixel-to-robot coordinate projection, and motion strategy selection based on detected pen angle (Figure 1). The color classifier labels each pen as blue, red, green, or grayscale by sampling the hue, saturation, and value (HSV) of pixels within the bounding box and selecting the dominant hue.

\hypertarget{hardware-setup}{%
\subsection{\texorpdfstring{\textbf{Hardware
Setup}}{Hardware Setup}}\label{hardware-setup}}

\begin{center}
\includegraphics[width=6.3in,height=4.725in]{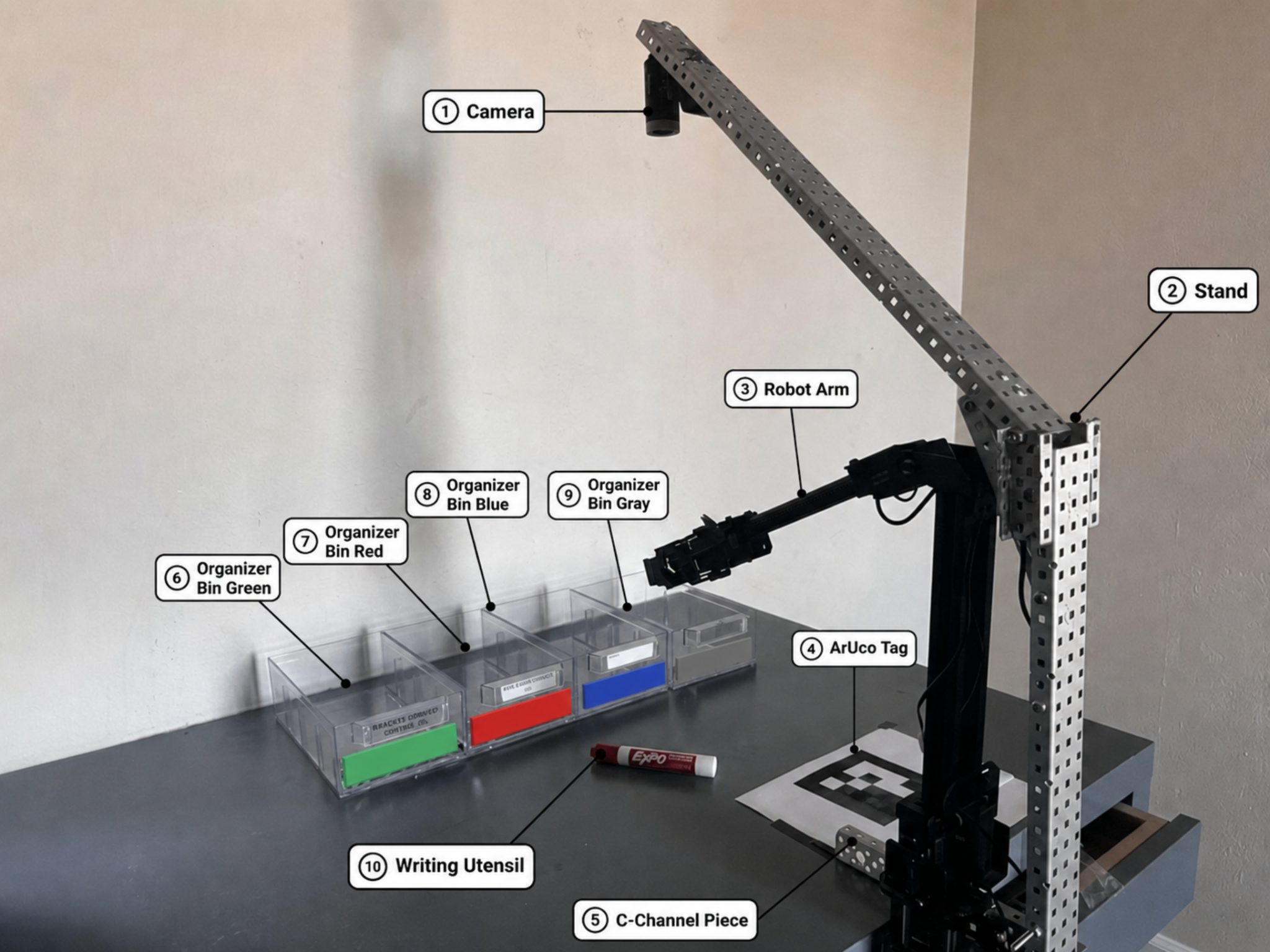}
\end{center}

Figure 1. Hardware setup, with components labeled (1) to (10). A 4-DoF robotic arm and camera mount over a painted wooden tabletop workspace with four color-coded sorting bins arranged in a row. An ArUco tag is placed on
the workspace to calibrate the extrinsics before beginning operations,
helping map camera detections to the robot frame during operation. Each
numbered component is identified in the list below.

\begin{quote}
1. Camera: Microsoft LifeCam Cinema (720p, widescreen), fixed overhead
for a top-down view of the workspace.

2. Stand: a custom frame of aluminum C-channels and screws that holds
the camera above the workspace.

3. Robot Arm: RoArm-M2-S (4 DoF) Waveshare arm with a serial JavaScript
Object Notation (JSON) command interface; it performs all grasping and
placement. Capabilities, wiring, and ranges are as specified on the
manufacturer's product page and wiki documentation {[}1, 2{]}.

4. ArUco Tag: fiducial marker for extrinsic calibration of the
camera-to-robot transform.

5. C-Channel Piece: aluminum extrusion that holds the ArUco tag at a
fixed, repeatable position.

6. Organizer Bin (Green): destination bin for utensils classified as
green.

7. Organizer Bin (Red): destination bin for utensils classified as red.

8. Organizer Bin (Blue): destination bin for utensils classified as
blue.

9. Organizer Bin (Gray): destination bin for utensils classified as
grayscale, meaning colors that do not substantially fall in the spectrum
of blue, red, or green.

10. Writing Utensil: the pen or marker to be detected, picked up, and
sorted.
\end{quote}

\hypertarget{end-effector}{%
\subsubsection{\texorpdfstring{{End-effector}}{End-effector}}\label{end-effector}}

\begin{center}
\includegraphics[width=2.91667in,height=3.84996in]{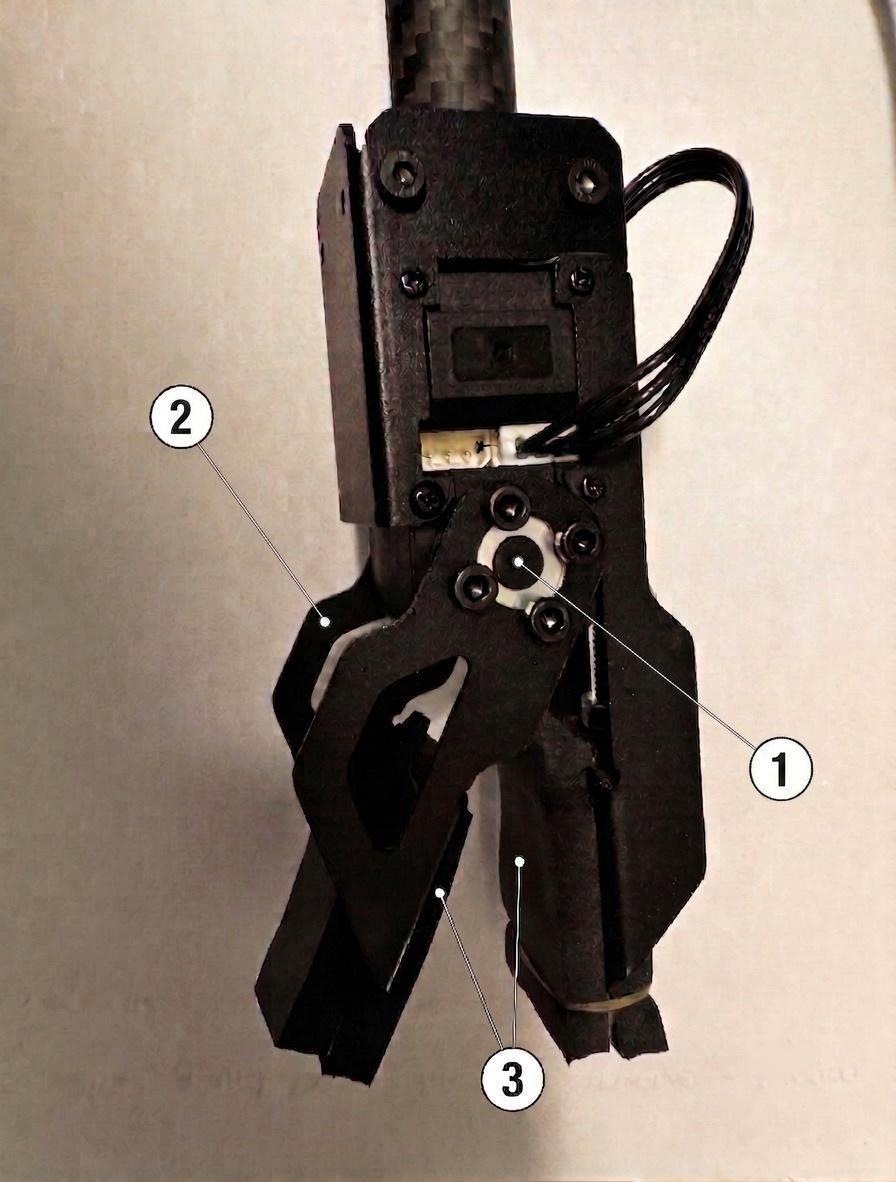}
\end{center}

Figure 2. The end-effector: (1) the pivot joint that drives the jaws,
(2) the single side-opening finger (the opposite jaw is fixed), and (3)
the foam-padded grasping tips.

The gripper is a single rotating jaw joint (Figure 2), and its angle is commanded in radians over about 118 degrees of travel. An angle of 1.08 holds the jaws fully open, and 3.14 holds them fully closed. Every grasp closes the joint to a fixed 3.0, just short of fully shut, then opens it back to 2.0 at the bin. The sweeps hold the joint at 2.95, so the jaws are nearly closed whenever the arm pushes a pen. The soft foam tips let it gently hold different pens without pressing so hard that it strains the motor.

\hypertarget{camera-calibration-intrinsics}{%
\subsection{\texorpdfstring{\textbf{Camera Calibration:}
Intrinsics}{Camera Calibration: Intrinsics}}\label{camera-calibration-intrinsics}}

Using a \(9 \times 6\) checkerboard with 209 views, the camera intrinsics \(K\), the 3×3 matrix containing the focal lengths \(f_{x},f_{y}\) and the principal point \(c_{x},c_{y}\), and distortion are estimated via Zhang's method [12]. The board had 9 by 6 inner corners with 22 mm squares, and the photographs were taken at 1280 by 720 pixels. The calibration gave focal lengths of fx = 985.5 px and fy = 984.5 px, a principal point at cx = 648.8 px and cy = 362.5 px, and distortion coefficients (k1, k2, p1, p2, k3) = (-0.0028, 0.0825, -0.0040, 0.0013, -0.2502). The root-mean-square (RMS) reprojection error of the calibration is 2.65 px. The focal lengths control the projection scale of the pixels, mapping the physical distances to the image coordinates. The principal point is the point where the optical axis intersects the image plane, typically near the image center. For each board pose \(i\) with corners \(X_{j}\) (the same known 3D corner positions in the checkerboard frame for all 209 views, measured in millimeters with z=0), corner j is observed at the pixel location \(u_{\text{ij}}\), modeled below:

\[
u_{\text{ij}} \approx K \cdot D(\Pi(\lbrack R_{i}\ |\ t_{i}\rbrack X_{j}))
\]

\(u_{\text{ij}}\) is the observed 2D pixel location of corner j in image
i, \(\Pi\) denotes the perspective projection (3D converted to 2D
location when divided by depth), \(D\) applies the lens distortion, K is the intrinsics being solved for, and
\(\lbrack R_{i}\ |\ t_{i}\rbrack\) transforms the checkerboard frame into the camera frame; pixel quantities are in pixels and checkerboard coordinates in millimeters. The letter D is reused in the Sweep Motion section for the sweep destination point; the two are unrelated. The total reprojection error, defined as the sum
of the squared distances between observed and predicted pixel locations
across all corners and views, is minimized to estimate K and the
distortion parameters.

\hypertarget{camera-calibration-extrinsics}{%
\subsection{\texorpdfstring{\textbf{Camera Calibration:}
Extrinsics}{Camera Calibration: Extrinsics}}\label{camera-calibration-extrinsics}}

A fixed workspace ArUco tag defines the world plane (Figure 3). Its
pose is estimated with estimatePoseSingleMarkers (OpenCV's marker-pose
function, which solves the Perspective-n-Point problem from the tag's
four detected corners). This returns a Rodrigues rotation vector and a
3×1 translation vector \(t\), and cv2.Rodrigues converts the rotation
vector into a 3×3 rotation matrix R. Together, R and t map a point from the tag frame into the camera frame, with t and the camera-frame point in meters, the unit OpenCV returns for a tag whose side is given in meters:

\[
X_{\text{cam}}\  = \ RX_{\text{tag}}\  + \ t,
\]

The system stores the last valid pose in Aruco/aruco\_reference.json. The tag itself is a 100 mm printed marker from OpenCV's 4 by 4 dictionary of 50 markers (DICT\_4X4\_50). The pose was re-estimated before each trial session, and in the stored poses the tag center lay 576 to 581 mm from the camera along the optical axis, so the camera-to-table distance is about 580 mm. The repeatability of the pixel-to-robot mapping was measured from the trial data (see Calibration and Detection Performance in the Results). The 100 mm tag maps back to 99.4 to 100.2 mm, and the mapping repeated to within 3.1 mm and 0.9 degrees within a session and to within 8.3 mm between sessions. No separate ruler measurement was made. The implementation uses OpenCV's 4.x ArUco tutorial [13] and the fiducial design of Garrido-Jurado et al. for reliable markers [14].

\begin{center}
\includegraphics[width=5.66667in,height=3.36346in]{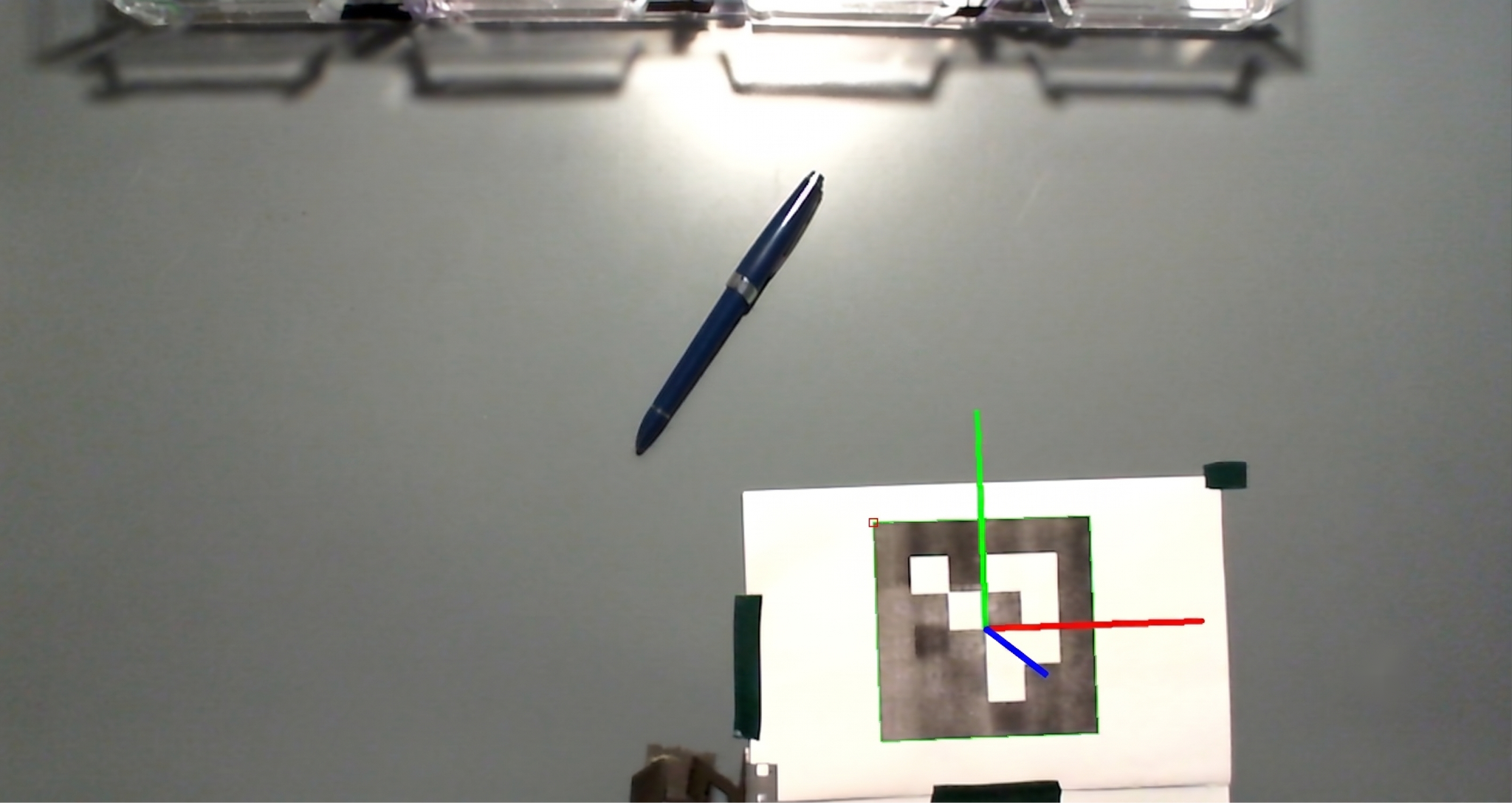}
\end{center}

Figure 3. Extrinsic calibration. The estimated pose of the workspace
ArUco tag is overlaid on the camera image. The colored axes mark the tag frame (red X, green Y, blue Z), and the green box marks the detected
tag. This pose, the rotation R and translation t, maps tag coordinates
into the camera frame.

\hypertarget{real-time-pen-detection}{%
\subsection{\texorpdfstring{\textbf{Real-Time Pen
Detection}}{Real-Time Pen Detection}}\label{real-time-pen-detection}}

A custom YOLO11n-OBB model [15] was trained on 540 images, using a single class labeled ``pen'': 210 captured in the workspace (30 per utensil across the seven writing utensils used in the trials) and 330 from an earlier Roboflow project of the first author, described below. The images were labeled with oriented bounding boxes in Roboflow [16] before training in Google Colaboratory [17]. The model (best.pt) runs on the live camera feed and outputs the oriented bounding box parameters \((c_{x},c_{y},w,h,\theta)\), where center coordinates are \({(c}_{x},c_{y})\), width (w), height (h) in pixels, and angle orientation (\(\theta)\) in radians. The OBB pipeline and formats follow the Ultralytics documentation on oriented bounding boxes [18].

The 330 online images come from the first author's earlier Roboflow project, Pens (version 2) [19]. It holds 128 source photographs. Of these, 118 were downloaded from the public Roboflow Universe dataset ``pen'' by the user egogor, which is released under the CC BY 4.0 license [20], and 10 were taken by the first author.

Roboflow made three augmented copies (flips, rotations, crops, shear, color shifts, blur, and noise) of each of the 101 source photographs in its training split, which gives 303 files. The other 27 went to validation and test unaugmented, so the 330 files are 303 copies plus 27 originals. Roboflow then split all 540 images into 362 training, 124 validation, and 54 test images, added no further augmentation, and stretched each image to 512 by 512 pixels.

Training used Ultralytics 8.4.62 [15] in Google Colaboratory. It started from the pretrained yolo11n-obb.pt weights and ran for 100 epochs at an image size of 512 pixels with a batch size of 16. The optimizer was set to automatic, and Ultralytics chose AdamW (Adam with decoupled weight decay) with an initial learning rate of 0.002, a momentum of 0.9, and a weight decay of 0.0005. The random seed was 0. The default training-time augmentation (mosaic, color jitter, translation, scaling, and horizontal flips) was left on.

The checkpoint used in all trials is best.pt from epoch 90. On the validation split, it reached a precision of 0.968 and a recall of 0.985. Its mean average precision (mAP) was 0.981 at an intersection-over-union (IoU) threshold of 0.50 (mAP50) and 0.718 when averaged over IoU thresholds from 0.50 to 0.95 (mAP50-95). Orientation error, measured as the difference between the detected box angle and the principal axis of the pen's pixel mask on the detector's 210 raw workspace photographs, had a median of 1.1 degrees and a 95th percentile of 3.5 degrees (see Calibration and Detection Performance in the Results). Inference speed was measured after the trials with the same weights and the same model call as the pipeline, as the wall-clock time of one call on a 1280 by 720 frame, averaged over 210 workspace photographs after 10 warm-up runs. On a 2-core Intel Xeon CPU at 2.10 GHz with no GPU, a different machine from the trial laptop but with the same Ultralytics 8.4.178 and PyTorch 2.14 that the laptop ran during the trials, it took 46.4 ms per frame on average (median 45.6 ms, range 34 to 62 ms), 21.5 frames per second. The loop rate during the sessions is given in the Results.

The split was made by image, not by utensil. The seven trial utensils therefore appear in the training images and can also appear in the validation images. These metrics do not show how the detector handles utensils it has not seen.

\hypertarget{filtering-the-view-and-on-screen-display}{%
\subsection{\texorpdfstring{\textbf{Filtering the View and On-Screen
Display}}{Filtering the View and On-Screen Display}}\label{filtering-the-view-and-on-screen-display}}

To avoid falsely detecting utensils in the storage bins and overhead
clutter in the camera frame, detections whose centers lie in the top
20\% of the frame are ignored:

\[
y_{\text{center}} < 0.2\, H.
\]

The user interface (UI) shades this strip as ignored (see the motion
step images in Figure 4). For accepted OBBs, the overlay shows the
oriented box and confidence, the short-edge midpoints (tips), and text
blocks with pixel/mm sizes, robot/tip coordinates, and penRadialAngle
(reported in radians and degrees).

The interface is built with OpenCV. The live camera feed is captured
with OpenCV's VideoCapture and shown in a HighGUI window, with keyboard
controls read through waitKey. All overlays use OpenCV drawing
functions: the ignored strip is a filled rectangle blended over the
frame with addWeighted, each accepted box is drawn with polylines, its
class and confidence label with putText on a filled rectangle, the two
short-edge midpoints with circle, and the size, coordinate, and
penRadialAngle readouts with putText. Color is read with the same
library by converting the box interior to HSV. In headless web mode, the
same OpenCV-drawn frame is streamed to a browser rather than a desktop
window. When the full pipeline is launched through full\_run.py, this
browser view runs in a compact desktop app-window (a Chrome app-mode
window served by a small local web server): during setup it provides
buttons to jog the arm onto the ArUco tag and confirm its position and
to capture the calibration photo, and during sorting it adds an AUTO
toggle, a manual trigger, and a stop button beside the live detection
feed.

\hypertarget{estimating-pen-color}{%
\subsection{\texorpdfstring{\textbf{Estimating Pen
Color}}{Estimating Pen Color}}\label{estimating-pen-color}}

Color is read from the pixels inside the detected box using a custom rule built on OpenCV's color conversions. The system fills the oriented box into a mask and slightly shrinks it to exclude border pixels, which avoids errors at the box's edges. It also drops pixels that are bright but washed out, since those are usually glare. It then checks how colorful the patch is overall, measured as the median chroma in the CIELAB (LAB) color space. If the patch is barely colorful, it is immediately labeled grayscale, without a hue check. Otherwise, the colored pixels are used to vote on whether the pen is blue, green, or red. If no color clearly wins, the system compares the average red, green, and blue levels and selects the strongest one if it clearly exceeds the other two, and otherwise the top hue vote stands.

\hypertarget{converting-pixels-to-robot-coordinates}{%
\subsection{\texorpdfstring{\textbf{Converting Pixels to Robot
Coordinates}}{Converting Pixels to Robot Coordinates}}\label{converting-pixels-to-robot-coordinates}}

This step turns a pen's location in the camera image into a real
position the arm can reach. The detector gives the pen's center as a
pixel (u, v), along with the angle of its oriented box. What the arm
needs instead is the same pen expressed in millimeters in the robot's
own coordinate frame. This is done in two stages: the pixel is first
mapped into a fixed real-world frame anchored to the ArUco tag on the
workspace, and that world position is then mapped into the robot's base
frame. The complication is that one camera cannot sense distance on its
own: a single pixel does not correspond to a single 3D point, but to a
whole ray leaving the camera, because every point along that line lands
on the same pixel. The depth is determined by one fact about the setup:
every pen lies flat on the table, which is in the same plane as the
ArUco tag. Following the pixel's ray until it meets that plane yields
the pen's real location, and since all pens share this one surface, each
is assigned the same z-coordinate. This height is fixed once, in a
reposition step at the start of a run, by jogging the arm until its
gripper rests on the center of the ArUco tag and saving the robot's own
coordinates there, which records the tag's position in the robot's
frame, calibrating the commanded pick depth to the physical table
surface. When it later picks a pen, this height is the depth its gripper
descends to before closing, so every pen is grasped at the same level.

First, from pixel to world coordinates. The procedure begins with the
OBB center \((u,v)\), the pen's pixel location, and combines three
quantities from calibration, each expressed in the camera's frame:

\(r = K^{- 1}\lbrack u\ v\ 1\rbrack^{\top}\) is the camera ray, the line
of sight pointing out through that pixel.

\(n = R\lbrack:,2\rbrack\) points straight up from the table,
perpendicular to the workspace plane; it is read from the tag's rotation
R, which is estimated from the tag's pose.

\(X_{0} = t\) is a point already known to lie on the plane: the ArUco
tag's own center, whose position in the camera frame is the translation
t recovered when the tag's pose was estimated.

Intersecting the ray with that plane pins down the pen:

\[
\lambda = \frac{n^{\top}X_{0}}{n^{\top}r},\quad X_{\text{cam}} = \lambda r,\quad X_{\text{tag}} = R^{\top}(X_{\text{cam}} - X_{0}).
\]

Here, λ is how far along the ray the table plane lies,
\(X_{\text{cam}}\) is the pen's 3D point in the camera frame, and
\(X_{\text{tag}}\) re-expresses that same point in the ArUco tag's frame, which serves as the world reference. The subscripts cam, tag, and robot name the frame in which a point is expressed; camera- and tag-frame points are in meters and robot-frame points in millimeters, the tag-frame point being scaled from meters to millimeters before the rotation below.

Second, from world to robot coordinates. This needs the tag's location
in the robot's own coordinates, which is exactly what the reposition
step above recorded. A tag-frame point then reaches the robot's frame in
two parts: a rotation that lines up the two frames' axes, then a shift
to the robot's origin:

\[
X_{\text{robot}} = R_{z}(\varphi)\,(X_{\text{tag}}) + p_{\text{offset}}
\]

The first step rotates the point about the vertical (z) axis, spinning
its x and y coordinates by the angle φ while leaving its height
unchanged, thereby swinging the tag's axes around until they line up
with the robot's. Written out \(\ R_{z}\)\((\varphi)\) is the matrix:

\[
\begin{bmatrix}
\cos\varphi & -\sin\varphi & 0 \\
\sin\varphi & \cos\varphi & 0 \\
0 & 0 & 1
\end{bmatrix}
\]

\(\varphi\  = \ 270{^\circ}\) is fixed by how the tag is mounted relative to the robot; it is a constant of the setup, not something estimated at runtime, and the matrix above is evaluated with the angle in radians (\(3\pi/2\), which is 270 degrees). \(p_{\text{offset}}\) then slides the rotated point to the robot's origin; it is the tag center expressed in robot coordinates, in millimeters. It is the robot's position saved during the positioning step before we scan the tag. In this step, we position the robot's end-effector at the tag's center and record the robot's coordinates at that point. The z position we collect there will also include the table level to which the arm descends during grasping.

The OBB's two short-edge midpoints are also projected using this method
to analyze tip geometry.

Orientation is computed in the world frame. The two short-edge midpoints
are back-projected through the same ray-plane procedure to robot
coordinates, and the pen's axis and its angle to the robot base are
computed from those projected points.

\hypertarget{motion-planning-and-arm-control}{%
\subsection{\texorpdfstring{\textbf{Motion Planning and Arm
Control}}{Motion Planning and Arm Control}}\label{motion-planning-and-arm-control}}

\hypertarget{motion-planning-overview}{%
\subsubsection{\texorpdfstring{{Motion Planning
Overview}}{Motion Planning Overview}}\label{motion-planning-overview}}

A detection is accepted only when the model is at least 70\% confident
it found a pen. In AUTO mode, detections are evaluated continuously. While the arm is moving, additional detections do not start overlapping motions. After the current motion completes, the system reevaluates the live camera feed and can trigger a subsequent motion for a currently detected pen. The detections do not drive the arm directly:
each queued detection launches one pick-and-place motion, and the arm is
driven by JSON commands sent over the serial link, each giving a target
position and a gripper state.

For each pen, the arm then runs the same sequence: it unfolds, stages
above the workspace, approaches the pen, descends, grips, lifts, routes
to the matching color bin, and returns home. Every command is checked
against the workspace limits {[}2{]}: the radial reach from the base
must be between 80 mm and 500 mm, and the height must be between -100 mm
and 450 mm; anything outside these limits is rejected and logged,
protecting the motors from strain or overheating.

\hypertarget{choosing-the-motion-by-pen-angle}{%
\subsubsection{\texorpdfstring{{Choosing the Motion by Pen
Angle}}{Choosing the Motion by Pen Angle}}\label{choosing-the-motion-by-pen-angle}}

The arm cannot turn its gripper to meet a pen at an arbitrary angle: the
direction it approaches from is fixed by where the pen sits on the
table. It can therefore grasp a pen directly only when the pen is
already close to that approach direction. In these cases, the system
measures how far each pen is turned away from it, the pen-radial
misalignment angle, and uses that one angle to choose between a direct
grasp and a corrective sweep. The angle is built up as follows.

Let \(C = (x,y,z)\) be the OBB center in robot coordinates and
\(T_{1},T_{2}\) the short-edge midpoints (projected to robot space), all in millimeters. In what follows a hat marks a unit vector, the subscript xy keeps the horizontal (x, y) components of a point, and the angle brackets denote the dot product.

Tip selection: Project each tip-center vector onto the radial line to
the origin. The system prefers tips whose projections lie toward the
origin, particularly those whose points are closest to the origin (to
prevent robot overreach). The selected tip is \(T^{\star}\). Note that
the system simply determines the short-edge midpoints based on geometric
criteria, not distinguishing between the physical tip and end of the
pen.

Pen-radial misalignment. The pen's own direction, from its center to the
chosen tip, is compared with the radial direction from the center toward
the robot's base. Writing the first as p and the second as r, they are:

\[
\widehat{p} = ((T^{\star} - C)_{\text{xy}})/ \parallel (T^{\star} - C)_{\text{xy}} \parallel \qquad \widehat{r} = \frac{\left( - x,\  - y \right)}{\parallel x,\ y \parallel}
\]

The misalignment is the angle between them, stored in radians as pen\_radial\_angle\_rad and reported in degrees as penRadialAngle:

\[
arccos(\langle\widehat{p},\widehat{r}\rangle)
\]

Because the chosen tip always points toward the base, this angle stays between 0 and 90 degrees, so no sign flip or absolute value is needed. The system then selects the motion with a single threshold:

\[
STANDARD\ if\ penRadialAngle < 45^{\circ},\quad COMPLEX\ otherwise.
\]

\hypertarget{standard-grasp}{%
\subsubsection{\texorpdfstring{{Standard
Grasp}}{Standard Grasp}}\label{standard-grasp}}

If the gripper aimed at the pen's centerline, the robot's right-opening
finger could flick the pen, so the standard instead grasps at a point 10
mm to the left of the pen center, perpendicular to the pen axis (Figure
4b defines this geometry). The pen axis \(\widehat{p}\) runs through the
tips \(T_{1},T_{2}\). Its leftward perpendicular, a planar unit vector, and the two candidate grasp points, which keep the height of the center because the semicolon appends a zero z component to the offset direction, are:

\[
{\widehat{n}}_{\text{\!L}} = ( - {\widehat{p}}_{y},{\widehat{p}}_{x}) \qquad C_{\pm} = C \pm d_{\perp}\lbrack{\widehat{n}}_{\text{\!L}};0\rbrack \qquad d_{\perp} = 10\ \text{mm}
\]

The candidate with the larger Y, the more leftward point, and the filled square in Figure 4b, is kept so that the right finger clears the pen on the way down.

The arm then runs a fixed command sequence. It stages clear of the workspace and lifts to travel height, moves over the grasp point, and lowers in two stages, first to a hover about 110 mm above the pen and then to the pick height z (see Converting Pixels to Robot Coordinates), where it closes the gripper partway to grip the pen. It lifts again, carries the pen to the bin for its color (all four bins sit at the same forward distance, x = 480 mm), lowers, releases, and returns home.

\begin{center}
\includegraphics[width=6.3in,height=6.55714in]{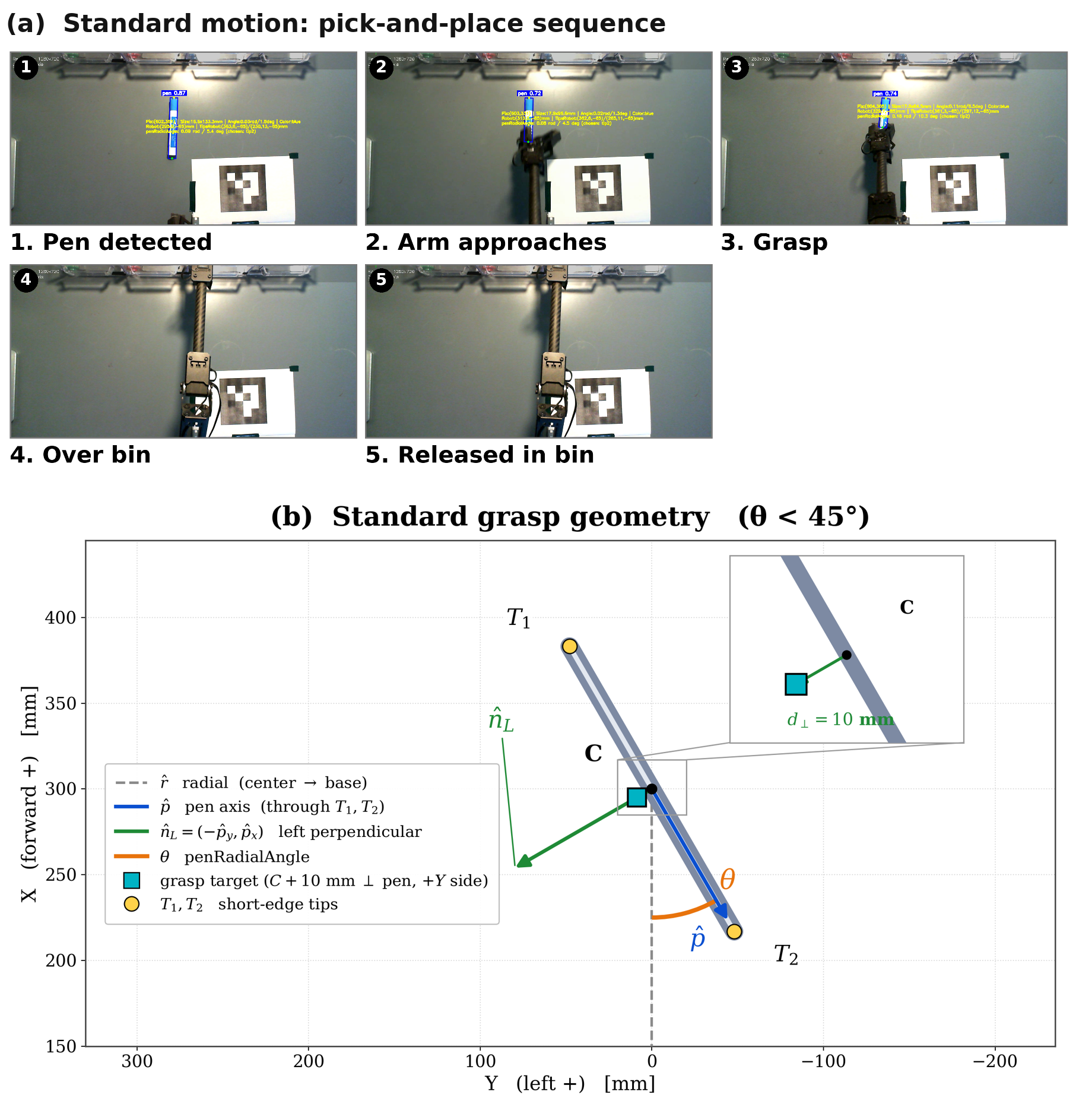}
\end{center}

Figure 4. Standard grasp. Panel (a) is a sequence of runtime camera frames with the YOLO11n-OBB oriented box in blue, the short-edge tips as small dots (green for the chosen tip, yellow for the other), and the measured penRadialAngle overlaid. Panel (b) is a schematic of the grasp geometry in robot coordinates, axes in millimeters, marking the OBB center, the pen axis, the radial line from the center to the robot base, and penRadialAngle, the angle between them. When that angle is below 45 degrees, the arm grasps directly at the cyan square, a point 10 mm from the center along the perpendicular, so the right finger clears the pen instead of flicking it.

\hypertarget{sweep-motion-for-steep-angles}{%
\subsubsection{\texorpdfstring{{Sweep Motion for Steep
Angles}}{Sweep Motion for Steep Angles}}\label{sweep-motion-for-steep-angles}}

When a pen is tilted more than 45 degrees, it is too far from the
gripper's fixed approach direction to grasp, so the arm reorients it
instead of picking it up. With its gripper nearly closed (2.95, see End-effector), the arm presses against the pen and drags it a short way toward the robot's base.

For a pen past 45 degrees, a local frame is built at the pen center, with one axis, the unit vector \(\hat{x}'\), pointing along the radial line toward the origin (Figure 5b); A and D below are planar points in millimeters. The approach point A is a point along the pen near the chosen tip (the orange circle A in Figure 5b), biased toward it:

\[
A = 0.25\, C_{\text{xy}} + 0.75\, T_{\text{xy}}^{\star}
\]

The destination D, the green marker in Figure 5b, then lies one
approach-distance toward the origin:

\[
D = C_{\text{xy}} + d_{\text{CA}}{\widehat{x}}^{\prime} \qquad d_{\text{CA}} = \parallel A - C_{\text{xy}} \parallel
\]

A sweep is initialized starting past \(A\) (the sweep start I in Figure
5b) and marches in approximately 10 mm steps toward \(D\) along the red
path, stopping 10 mm early. If more than seven waypoints are produced,
the last four are dropped to avoid overshoot. The sweep start lies on
the line through D and A, extended beyond A, with the distance from D to
I equal to 2∥A−D∥ + 10 mm (about ∥A−D∥ + 10 mm past A).

The waypoints stay at the pick height \(z\). This is the \(z\) that is
set during the calibration of the workspace with ArUco extrinsic
calibration. This height is simply the robot's own height, recorded at
the tag during setup.

\textbf{Choice of parameters:} The five values were set from the geometry of the gripper and the utensils and were not tuned. At 45 degrees a pen's axis is as far from the jaws' closing direction as from the line across it, so below that angle the jaws close across the pen and above it they do not. The 10 mm offset keeps the right finger from coming down on the pen. The 10 mm step keeps each serial command to a small move, and dropping the last four waypoints of a path longer than seven keeps a long sweep from pushing the pen past the radial line. The approach point \(A = 0.25\,C + 0.75\,T\) sits near the tip because a push far from the center of friction turns a pen rather than sliding it [4]. None of these values was varied in the trials, so sensitivity to them was not measured.

\begin{center}
\includegraphics[width=6.3in,height=6.94633in]{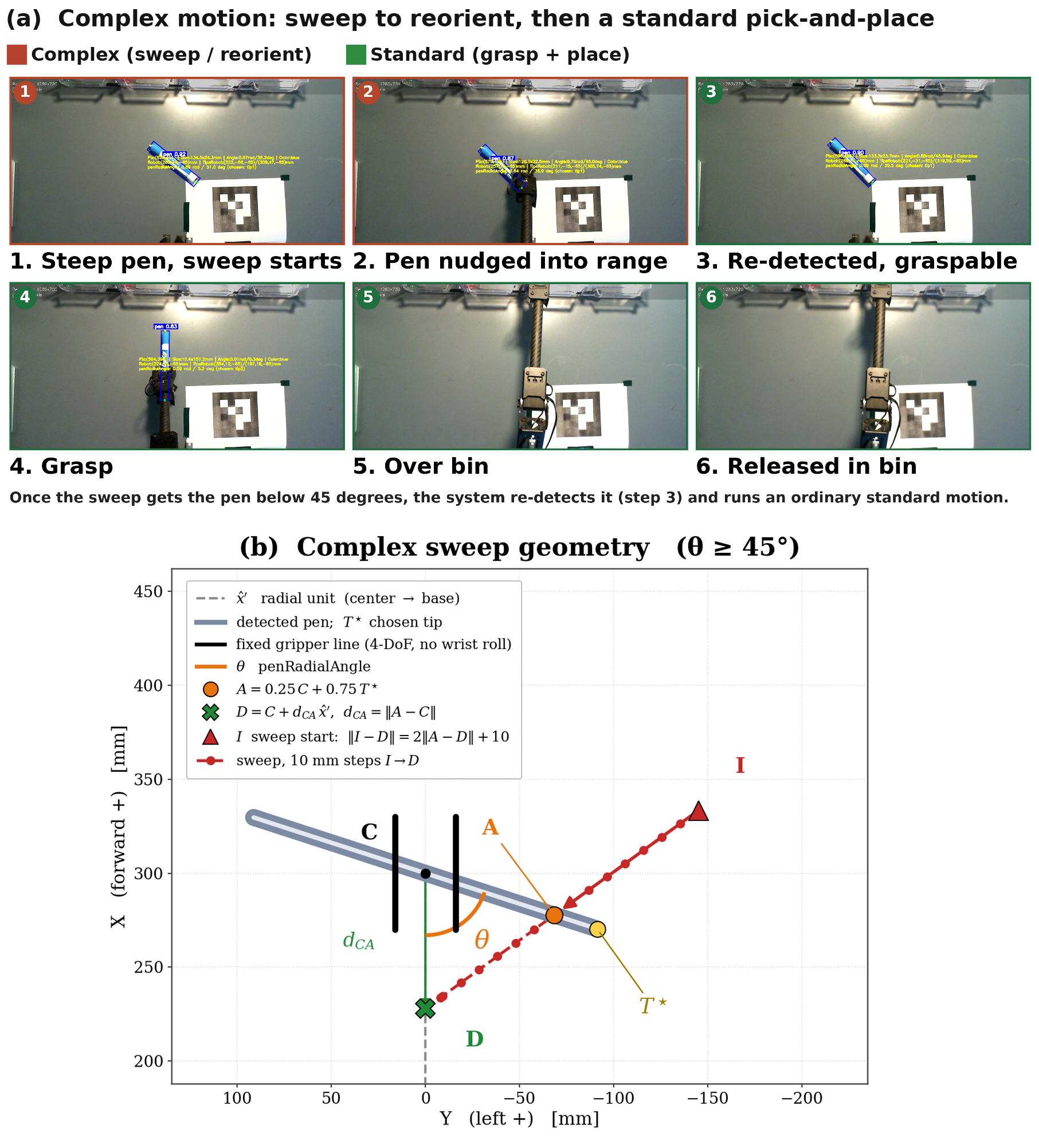}
\end{center}

Figure 5. Complex sweep. Panel (a) is a sequence of runtime camera frames of a pen past 45 degrees, with the YOLO11n-OBB oriented box in blue and the measured penRadialAngle in the overlay. Panel (b) is a schematic of the sweep geometry in robot coordinates, axes in millimeters. Because the 4-DoF arm has no wrist roll, the gripper has one fixed closing orientation along the radial line, so a pen lying across it cannot be grasped directly. With the jaws nearly closed (2.95), the arm drags the pen from the sweep start toward the base in 10 mm steps, using the approach point A and the destination D built from the chosen tip, which usually lowers the angle below 45 degrees, so a following pass can grasp it. The chosen tip blinks in the live Matplotlib plot, so in a still frame of that plot, its marker may appear faint, even though both tips are detected.

\hypertarget{sorting-pens-by-color}{%
\subsubsection{\texorpdfstring{{Sorting Pens by
Color}}{Sorting Pens by Color}}\label{sorting-pens-by-color}}

Final color labels are \{blue, red, green, grayscale\}. The routing to the color boxes after pickup uses \(y = \{ - 70, + 70, + 140, - 140\}\) mm, respectively. The \(x\) location is a set-forward distance from the origin, since all boxes share that \(x\).

\hypertarget{full-operating-sequence}{%
\subsubsection{\texorpdfstring{{Full Operating
Sequence}}{Full Operating Sequence}}\label{full-operating-sequence}}

full\_run.py is the orchestration script that runs the whole pipeline
end-to-end. It first guides the operator through jogging the arm onto
the ArUco tag and confirming its position, then capturing the
calibration photo. It refreshes the ArUco pose in
Aruco/aruco\_reference.json and launches camera\_stream.py to detect and
sort pens. It also keeps four calibration artifacts saved on disk:

\(K\), the camera's intrinsic matrix;

\(\text{dist}\), the lens distortion coefficients;

\(\text{rvec}\), the ArUco tag's rotation, stored as a Rodrigues vector;
and

\(\text{tvec}\), the ArUco tag's translation.

Keeping these on disk means that camera\_stream.py reads them once at
startup, so the same calibration is used for every detection in a session.

\hypertarget{visualization-and-debugging}{%
\subsubsection{\texorpdfstring{{Visualization and
Debugging}}{Visualization and Debugging}}\label{visualization-and-debugging}}

While OpenCV draws the live camera overlay, a parallel Matplotlib plot reprojects the same geometry into robot coordinates with an equal aspect ratio. It plots the box center \(C\), the two short-edge tips \(T_{1},T_{2}\), and the offset grasp target \(C_{+}\), placed 10 mm to the side of the center so the right finger does not catch the pen.

For a complex motion, the plot also shows the sweep path, namely the
destination \(D\) toward the origin, the sweep start \(I\), and the
intermediate waypoints \(P_{k}\), the points along the linear sweep
path.

Each run saves timestamped PNGs of the annotated camera view.

\hypertarget{logging-and-reproducibility}{%
\subsection{\texorpdfstring{\textbf{Logging and
Reproducibility}}{Logging and Reproducibility}}\label{logging-and-reproducibility}}

Each session writes to a temporary folder under ResearchDataset/. On
termination, the system releases the video writer, prints the duration,
assigns the next log index, and automatically renames the folder. All
output is logged to the console and to log.txt, with a thread-safe
wrapper that flushes promptly and detaches cleanly on shutdown. The
first accepted frame initializes the MP4 writer at webcam FPS (default
30); key events triggered are synchronized with annotated camera stream
frames.

\hypertarget{results}{%
\section{\texorpdfstring{\textbf{RESULTS}}{RESULTS}}\label{results}}

\hypertarget{experiments}{%
\subsection{\texorpdfstring{\textbf{Experiments}}{Experiments}}\label{experiments}}

Trials were run one writing utensil at a time on the tabletop workspace: a wooden board with a smooth coat of light gray paint. For each utensil, the operator placed the pen by hand at starting angles stepped through nine 10-degree ranges from 0 to 90 degrees relative to the radial approach direction, in roughly increasing order rather than randomized, with one to seven trials per range. Detections were triggered manually so that each trial handled one pen from first detection to final placement, giving 23 to 31 trials per utensil and 196 trials in total. Seven utensils spanning a range of colors and shapes were tested (Figure 6). For every motion, the system logged the detected color, the initial penRadialAngle, the motion type, and the number of sweep waypoints (n\_sweeps); these logs underlie Figure 7 and Table 1.

All seven sessions were run on a single evening, one utensil per session. The only light was one fixed desk lamp, and the camera ran at 1280 by 720 pixels with autofocus off and automatic exposure. The operator placed the utensil by hand, triggered each detection from the web interface and reset the utensil after each trial. Every triggered detection ran its full motion sequence and was logged, failed trials included, and none was removed afterwards.

\begin{center}
\includegraphics[width=4.89063in,height=3.2369in]{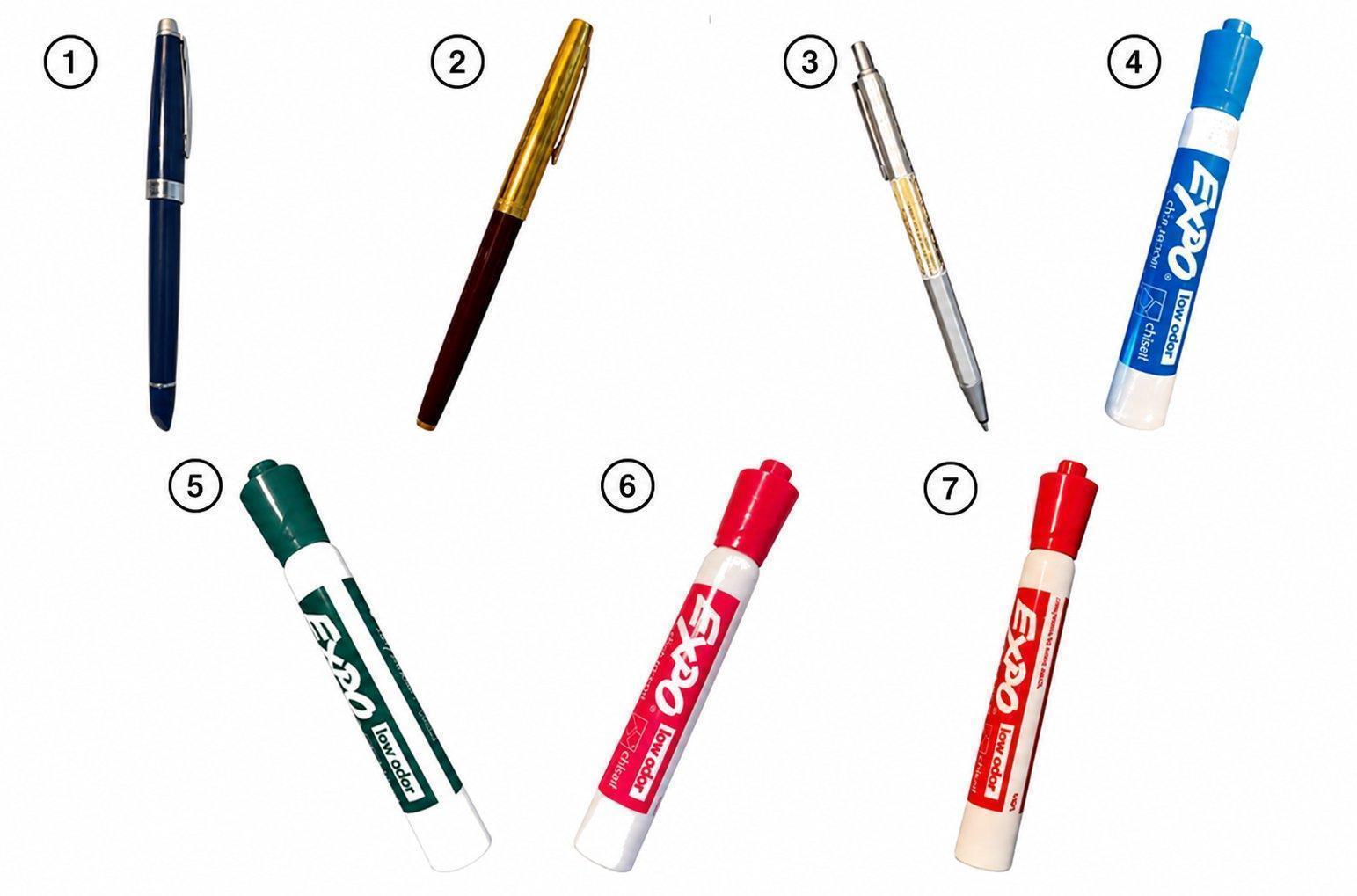}
\end{center}

Figure 6. The seven writing utensils used in the experimental trials,
labeled (1) to (7): (1) a navy pen, (2) a gold-and-maroon pen, (3) a
gray pen, and Expo dry-erase markers in (4) blue, (5) green, (6) pink, and (7) red.

\begin{center}
\includegraphics[width=6.3in]{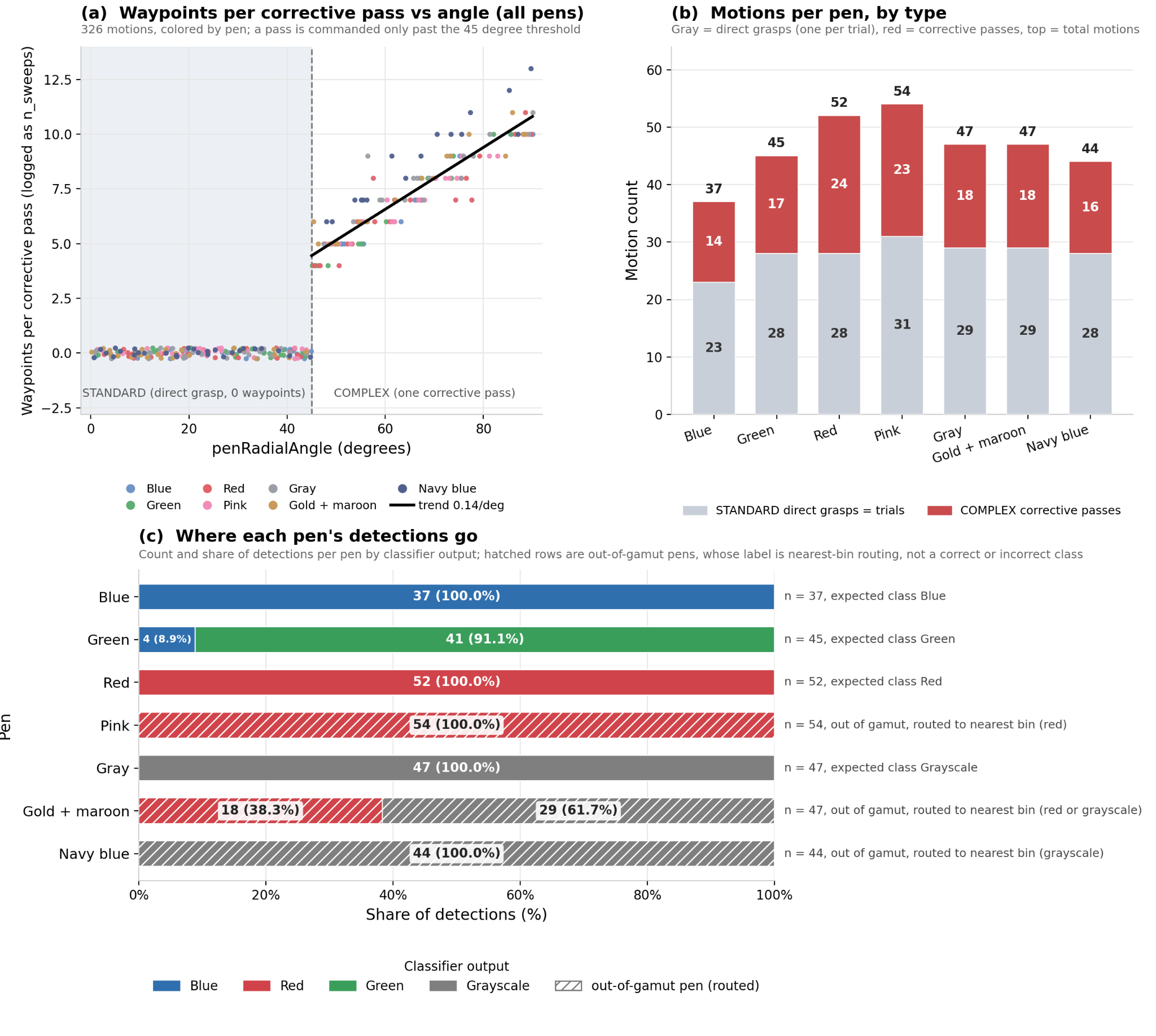}
\end{center}

Figure 7. Corrective sweeps, motion counts, and detected colors across the seven pens. Panel (a) plots the number of 10 mm sweep waypoints per corrective pass (logged as n\_sweeps) against penRadialAngle for every motion, zero below 45 degrees, and rising steadily above it; n\_sweeps counts waypoints within one pass, not complete sweep passes. Panel (b) shows the logged motions per pen: the gray segment and its number are the direct grasps, one per trial, the red segment and its number are the corrective sweep passes, and the number above each bar is the total. Panel (c) shows the detected color per pen as the count and share of detections assigned to each color bin; the four in-gamut pens have an expected class, and the hatched rows are the three out-of-gamut pens, whose label is nearest-bin routing rather than a correct or incorrect classification.

\begin{center}
\footnotesize\setlength{\tabcolsep}{2.2pt}\renewcommand{\arraystretch}{1.1}
\begin{tabular}{>{\raggedright\arraybackslash}p{1.05in}>{\raggedright\arraybackslash}p{0.47in}>{\raggedright\arraybackslash}p{0.47in}>{\raggedright\arraybackslash}p{0.47in}>{\raggedright\arraybackslash}p{0.47in}>{\raggedright\arraybackslash}p{0.47in}>{\raggedright\arraybackslash}p{0.47in}>{\raggedright\arraybackslash}p{0.47in}>{\raggedright\arraybackslash}p{0.47in}>{\raggedright\arraybackslash}p{0.47in}>{\raggedright\arraybackslash}p{0.47in}}
\hline
\textbf{Utensil} & \textbf{Trials} & \textbf{Passes} & \textbf{Grasped and placed} & \textbf{In correct bin} & \textbf{Grasp failures} & \textbf{Routing failures} & \textbf{Success rate, \%} & \textbf{95\% CI low} & \textbf{95\% CI high} & \textbf{Correct color label, \%} \\
\hline
Blue marker & 23 & 14 & 22 & 22 & 1 & 0 & 95.7 & 79.0 & 99.2 & 100.0 \\
Green marker & 28 & 17 & 28 & 27 & 0 & 1 & 96.4 & 82.3 & 99.4 & 91.1 \\
Red marker & 28 & 24 & 28 & 28 & 0 & 0 & 100.0 & 87.9 & 100.0 & 100.0 \\
Pink marker & 31 & 23 & 31 & 31 & 0 & 0 & 100.0 & 89.0 & 100.0 & n/a \\
Gray pen & 29 & 18 & 29 & 29 & 0 & 0 & 100.0 & 88.3 & 100.0 & 100.0 \\
Gold-and-maroon pen & 29 & 18 & 24 & 24 & 5 & 0 & 82.8 & 65.5 & 92.4 & n/a \\
Navy pen & 28 & 16 & 15 & 15 & 13 & 0 & 53.6 & 35.8 & 70.5 & n/a \\
\hline
\textbf{All seven} & \textbf{196} & \textbf{130} & \textbf{177} & \textbf{176} & \textbf{19} & \textbf{1} & \textbf{89.8} & \textbf{84.8} & \textbf{93.3} & \textbf{n/a} \\
\hline
\end{tabular}

\vspace{6pt}

\begin{tabular}{>{\raggedright\arraybackslash}p{1.35in}>{\raggedright\arraybackslash}p{0.44in}>{\raggedright\arraybackslash}p{0.44in}>{\raggedright\arraybackslash}p{0.44in}>{\raggedright\arraybackslash}p{0.44in}>{\raggedright\arraybackslash}p{0.44in}>{\raggedright\arraybackslash}p{0.44in}>{\raggedright\arraybackslash}p{0.44in}>{\raggedright\arraybackslash}p{0.44in}>{\raggedright\arraybackslash}p{0.44in}>{\raggedright\arraybackslash}p{0.44in}}
\hline
\textbf{Starting range, degrees} & \textbf{0 to 10} & \textbf{10 to 20} & \textbf{20 to 30} & \textbf{30 to 40} & \textbf{40 to 50} & \textbf{50 to 60} & \textbf{60 to 70} & \textbf{70 to 80} & \textbf{80 to 90} & \textbf{All ranges} \\
\hline
Blue marker & 3 & 5 & 2 & 4 & 2 & 2 & 2 & 1 & 2 & 23 \\
Green marker & 3 & 3 & 5 & 4 & 4 & 2 & 2 & 2 & 3 & 28 \\
Red marker & 4 & 3 & 1 & 2 & 3 & 4 & 4 & 4 & 3 & 28 \\
Pink marker & 3 & 3 & 3 & 3 & 4 & 3 & 4 & 5 & 3 & 31 \\
Gray pen & 3 & 5 & 2 & 1 & 3 & 4 & 5 & 3 & 3 & 29 \\
Gold-and-maroon pen & 2 & 7 & 1 & 3 & 3 & 4 & 3 & 3 & 3 & 29 \\
Navy pen & 3 & 3 & 3 & 3 & 3 & 4 & 3 & 3 & 3 & 28 \\
\hline
\textbf{Total trials} & \textbf{21} & \textbf{29} & \textbf{17} & \textbf{20} & \textbf{22} & \textbf{23} & \textbf{23} & \textbf{21} & \textbf{20} & \textbf{196} \\
\textbf{Total passes} & \textbf{0} & \textbf{0} & \textbf{0} & \textbf{0} & \textbf{12} & \textbf{26} & \textbf{32} & \textbf{27} & \textbf{33} & \textbf{130} \\
\textbf{Grasped and placed} & \textbf{21} & \textbf{27} & \textbf{16} & \textbf{19} & \textbf{20} & \textbf{19} & \textbf{20} & \textbf{17} & \textbf{18} & \textbf{177} \\
\textbf{In correct bin} & \textbf{21} & \textbf{27} & \textbf{16} & \textbf{19} & \textbf{20} & \textbf{19} & \textbf{20} & \textbf{16} & \textbf{18} & \textbf{176} \\
\textbf{Success rate, \%} & \textbf{100.0} & \textbf{93.1} & \textbf{94.1} & \textbf{95.0} & \textbf{90.9} & \textbf{82.6} & \textbf{87.0} & \textbf{76.2} & \textbf{90.0} & \textbf{89.8} \\
\textbf{95\% CI low} & \textbf{84.5} & \textbf{78.0} & \textbf{73.0} & \textbf{76.4} & \textbf{72.2} & \textbf{62.9} & \textbf{67.9} & \textbf{54.9} & \textbf{69.9} & \textbf{84.8} \\
\textbf{95\% CI high} & \textbf{100.0} & \textbf{98.1} & \textbf{99.0} & \textbf{99.1} & \textbf{97.5} & \textbf{93.0} & \textbf{95.5} & \textbf{89.4} & \textbf{97.2} & \textbf{93.3} \\
\hline
\end{tabular}
\end{center}

Table 1. Trial outcomes per utensil (upper part) and per 10-degree starting range (lower part; bold rows are totals over the seven utensils). Trials equal direct grasps, one per trial. Grasped and placed counts the trials in which the grasp held through to a bin, and in correct bin counts those that reached the bin for the utensil's color, or the routed bin for the three out-of-gamut utensils. Success rate is in correct bin divided by trials, with a 95\% Wilson confidence interval (CI). Correct color label is the share of detections labeled with the utensil's own color, for the four in-gamut utensils.

\hypertarget{quantitative-analysis}{%
\subsection{\texorpdfstring{\textbf{Quantitative
Analysis}}{Quantitative Analysis}}\label{quantitative-analysis}}

Each motion is scored by type. A standard motion succeeds when the arm grasps the pen and releases it into the bin for its color. The three out-of-gamut utensils have no bin of their own, so for them the bin that the color rule routed them to counts as the target, and whether the color label itself was right is judged separately, for the four in-gamut utensils only. A complex motion succeeds when its sweep leaves the pen below the 45-degree threshold, so that the standard grasp that follows can pick it up. The bin is chosen by the color label on the grasp detection. Whether a grasp held and which bin the utensil reached were scored from the session videos, and whether a sweep left the pen below the threshold is read from the next logged detection. The following terms are used in a fixed sense. A trial is one utensil handled from a triggered detection to the arm's return, and it always ends with exactly one direct grasp attempt, the pick-lift-carry-release motion. When the trial starts beyond 45 degrees, one or more corrective sweep passes come first, each a single push of the utensil from the sweep start point toward its end point, made of roughly 10 mm steps called waypoints. A detection is one accepted detector output, with a confidence of at least 0.7, that starts a motion, and a logged motion is either one direct grasp or one sweep pass, which is why the 326 logged motions are 196 grasps plus 130 passes.

In total, 326 motions were logged across seven pens: 196 direct grasps and 130 corrective sweep passes, divided at the 45-degree threshold. Because every trial ends with one direct grasp, the 196 direct grasps correspond to 196 trials, and 98 of those began beyond 45 degrees and needed at least one corrective pass. 176 of the 196 trials (89.8\%) ended with the utensil in the bin for its color. Of the 20 that did not, 19 were grasps that slipped or missed (13 on the navy pen, 5 on the gold-and-maroon pen and 1 on the blue marker), and the last was a green marker that the color rule labeled blue and the arm placed in the blue bin. By motion type, 92 of the 98 direct-grasp trials (93.9\%) and 84 of the 98 sweep-then-grasp trials (85.7\%) ended in the correct bin. Table 1 breaks these counts down by utensil and by starting-angle range. Past the threshold, the commanded sweep path increased linearly with angle (0.141 waypoints per degree, r = 0.92), consistent with the sweep geometry (Figure 7), and most initially misaligned pens were corrected in a single sweep pass (75 of 98), with steeper pens requiring two passes in 15 trials, three in 7, and four in 1. The largest angle at which a direct grasp ran was 44.93 degrees and the smallest at which a sweep ran was 45.07 degrees. Each corrective sweep reduced penRadialAngle by about 33 degrees on average, and 98 of the 130 corrective sweep passes (75\%) left the pen below the 45-degree grasp threshold on the following detection; color was correct for at least 91\% of detections on the four in-gamut pens (see Figure 7c and Table 1); the remaining three fell outside the four-bin color gamut and mapped systematically to the nearest bin (pink to red, navy to grayscale), with only the gold-and-maroon pen splitting between red and grayscale under glare. The navy pen's dark surface reads as barely colorful (low chroma), so the classifier labeled it grayscale before its blue hue was ever checked.

\hypertarget{calibration-and-detection-performance}{%
\subsection{\texorpdfstring{\textbf{Calibration and Detection Performance}}{Calibration and Detection Performance}}\label{calibration-and-detection-performance}}

At the stored pose, one pixel covers about 0.59 mm on the table, so the 2.65 px reprojection error of the intrinsic calibration corresponds to about 1.6 mm. Passing the four detected tag corners through the pixel-to-robot mapping returns sides of 99.4 to 100.2 mm for the 100 mm tag, which checks the implementation rather than its accuracy, since the pose was fitted to those same corners. In the 326 trigger frames, its corners sat a median of 0.8 mm and at most 3.4 mm from their positions in that session's calibration photo, and re-estimating the pose from those frames would move mapped positions in the work area by at most 3.1 mm and the tag's yaw by at most 0.87 degrees. The calibration was redone before each session, and recomputing each session's logged pixel centers through the final session's calibration instead of its own moves them by 8.3 mm at most, which bounds how reproducibly the calibration was re-established and includes the hand-jogged tag position (spread 5 mm in x, 5 mm in y, and 4 mm in z).

The saved trial frames carry the detection overlay, so the pen axis was measured on the detector's 210 raw workspace photographs. On these 210 photographs, the long axis of the detected box agreed with the principal axis of the pen's pixel mask within a median of 1.1 degrees, 3.5 degrees at the 95th percentile and 8.0 degrees at most, with the largest differences from masks that took in the pen's shadow or the white sheet under the tag. These are the raw workspace photographs from which the detector's dataset was built, taken under the trial camera, so the figure is not an error against labeled ground truth on unseen images. A difference of the 95th-percentile size would change the motion category only for a pen within 3.5 degrees of the 45-degree threshold, which was the case for 15 of the 196 trials. Inference took 46.4 ms per 1280 by 720 frame (median 45.6 ms, range 34 to 62 ms over the same 210 photographs after 10 warm-up runs) on a 2-core Intel Xeon CPU at 2.10 GHz with no GPU, measured after the trials with the same weights and model call on a different machine from the trial laptop. During the sessions, the perception loop wrote every annotated frame to the session video, and the frame counts and running times of the seven session videos give 8.3 to 8.8 frames per second, so detection, color classification, drawing, and streaming together took at most about 120 ms per frame on the trial laptop, a pace the camera's frame rate under the single lamp may have set.

\hypertarget{qualitative-results}{%
\subsection{\texorpdfstring{\textbf{Qualitative Performance and Failure Analysis}}{Qualitative Performance and Failure Analysis}}\label{qualitative-results}}

Across varied pen orientations and colors, the system exhibited two consistent behaviors: (i) standard motions grasped the pen and placed it in the correct bin in 176 of 196 trials once the pen's axis was within 45 degrees of radial, and (ii) complex motions brought pens with higher misalignments below that threshold in 98 of 130 passes, via one or more short lateral sweeps before the grasp.

Observed sweeps showed that the arm reorients pens both angularly, along the pen's long axis, and positionally. While the navy and gold-and-maroon pens pivoted toward a more radial orientation, the Expo markers and the gray pen (Figure 6) tended to translate rather than rotate, shifting to a more graspable position. The operational process was open-loop: the arm presses and lifts without sensing the pen mid-drag, and the camera re-evaluates the angle after each pass to decide whether a standard grasp can run or another sweep is needed. A single pass sufficed for 75 of the 98 trials beyond 45 degrees, and 20 of the 23 multi-pass trials (up to four passes) began above 60 degrees, so steeper pens needed more passes on average, consistent with the pass counts in the quantitative analysis.

In standard motions, the 10 mm left-offset target was meant to avoid right-finger flick; 177 of the 196 grasps (90.3\%) held through transport, and 13 of the 19 that slipped or missed were on the navy pen. In the complex motions, the approach point is biased toward the selected tip so that the push turns the pen rather than sliding it.

Angle noise from the OBB detector was not measured against a protractor; the detection checks in the Results give the agreement of the detected angle with the visible pen axis and how many trials began near the 45-degree boundary. The region-of-interest (ROI) guard (top 20\% exclusion) avoided false positives from color-sorting bin rims and clutter overhead in the boxes caused by utensils inside.

The trials that began above 85 degrees needed the most corrective passes, 1.8 on average against 1.3 between 60 and 85 degrees and 1.1 between 45 and 60, but only 2 of the 20 failed trials began above 85 degrees. During high-glare frames, HSV filtering suppressed valid pixels, briefly delaying class and angle stabilization. On 13 trigger presses, at seven moments in four sessions, no detection reached the 0.7 confidence cutoff and nothing was sent; a later press found the pen each time. The 326 accepted detections had confidences between 0.70 and 0.92 with a median of 0.82; 132 were below 0.80, and 64 of those came from the gray and gold-and-maroon pens, which also caused 11 of the 13 empty presses. The 20 failed trials were 19 grasps that slipped or missed and one green marker that the color rule misrouted; sweep inefficiency showed up as extra passes, not as missed detections. Overshoot past the radial line was not measured, since penRadialAngle is unsigned.

The dual visualization (OpenCV overlay
+ Matplotlib reconstruction) made mismatches between tag pose and the robot frame apparent.

\hypertarget{discussion}{%
\section{\texorpdfstring{\textbf{DISCUSSION}}{DISCUSSION}}\label{discussion}}

\hypertarget{principal-findings}{%
\subsection{\texorpdfstring{\textbf{Principal Findings}}{Principal Findings}}\label{principal-findings}}

Within this setting, the push-to-grasp behavior comes from a single geometric rule on the detected angle. Of the 196 trials, 176 (89.8\%) ended with the utensil in the bin for its color, 92 of the 98 direct-grasp trials and 84 of the 98 that needed sweeps. Of the 20 that did not, 19 were grasps that slipped or missed and one was a color misroute. All 98 trials that began beyond 45 degrees became graspable within four passes, a single pass sufficed for 75 of them, and each corrective sweep reduced the misalignment by about 33 degrees on average.

\hypertarget{comparison-with-prior-work}{%
\subsection{\texorpdfstring{\textbf{Comparison With Prior Work}}{Comparison With Prior Work}}\label{comparison-with-prior-work}}

The sweep is a planar push in Mason's sense [4]. Whether the body turns or slides depends on where the push line passes relative to its center of friction, which is one reading of why the same command rotated the round pens and translated the flatter markers and the gray pen. Another untested reading is that the flatter bodies raise their rotational friction relative to their translational friction, so that a sweep slides them more than it turns them. Dogar and Srinivasa's planner found and executed a push-grasp in four of five real scenes, and in three of five once 2 cm of pose noise was added [6]. Zeng and his colleagues reached 83.3\% grasp success and 71.4\% task completion in clutter on a UR5 arm [7]. The 176 of 196 trials (89.8\%) here are not directly comparable to those numbers, since this setting is narrower, with one known utensil at a time on a clear surface, three pose degrees of freedom and a rigid gripper, and a push chosen by one fixed rule. On the sensing side, Cong's 4-DoF sorting arm keeps correcting from the image as it moves and needs no camera calibration [10]; the system here is look-then-move rather than visual servoing [9], calibrates once per session and plans each motion from a single detection in robot coordinates, which lets it place a push point beside the utensil but leaves slip and friction differences uncorrected.

\hypertarget{limitations}{%
\subsection{\texorpdfstring{\textbf{Limitations}}{Limitations}}\label{limitations}}

On the same smooth painted surface, pen surface finish and mass led to variability in friction, so identical sweep or nudge motions did not move all utensils by the same distance. Glare at times reduced both detection confidence and color consistency. Because extrinsics are estimated once and then reused in a session, small shifts of the camera mount or arm base accumulate into millimeter-scale pose errors. The motion policies run open-loop: the sweep or grasp is planned from a single OBB detection and then executed without live orientation feedback, so any small motion of the utensil after detection is not corrected in the same run.

The evaluation has limits of its own. No direct-grasp-only baseline was run on the same starting poses, so the effect of the sweeps is not measured against another policy, and the 326 logged motions are not independent samples, since the 130 sweep passes belong to 98 trials and all trials of a utensil share one session. All trials used the same seven utensils, which also appear in the detector's training images, every detection was triggered by hand, and everything ran under one camera on one tabletop. The accuracy of the detected orientation and of the pixel-to-robot mapping was not measured against a protractor or a ruler; the checks in the Results bound their repeatability from the logged data, and the calibration repeated between sessions only to within about 8 mm, so the success rates describe the whole system and do not establish the error of either component on its own. How many of the 98 steep trials a direct grasp alone would have handled, and how the outcome changes with other threshold, offset, step, truncation or weighting values, would need new robot trials.

The trials say nothing about clutter, several utensils at once, other surfaces, or other lighting. Another surface changes the friction that decides whether a sweep rotates or slides a utensil, so the pass counts would change with it. None of these cases was run, so the 89.8\% figure describes this setup only.

\hypertarget{future-directions}{%
\subsection{\texorpdfstring{\textbf{Future Directions}}{Future Directions}}\label{future-directions}}

A direct comparison is needed first. Repeating the same angle protocol on utensils held out of the training set, in randomized order, both with corrective sweeps and with sweeps disabled, would give a direct-grasp-only baseline and complete grasp-and-sort success rates, and the same runs would measure the orientation error and the pixel-to-robot error directly against a protractor and a ruler. The friction differences are a second line of study. Adaptive nudging policies that reason about contact could better handle them, for example, by choosing between a short nudge near the end of the utensil, which rotates it, and a longer sweep nearer its middle, which slides it, by modulating sweep distance based on observed motion or inferred contact force, or by estimating a utensil's friction from how far its first pass moved it. Trials on other surfaces and with hexagonal or clipped bodies would test how friction and contact geometry decide whether a sweep rotates or slides a utensil. Re-running the OBB detector between waypoints or using visual servoing on the pen axis would turn the open-loop sweeps into closed-loop behaviors that respond to slip as it happens, and re-estimating the tag pose from each frame would absorb drift. Expanding the training set with more images per pen type and color, additional lighting conditions, and harder viewpoints would improve detection robustness.

\hypertarget{conclusion}{%
\section{\texorpdfstring{\textbf{CONCLUSION}}{CONCLUSION}}\label{conclusion}}

This work investigated whether a low-cost 4-DoF arm that lacks wrist rotation can pick up and color-sort writing utensils using perception and motion planning, rather than additional mechanical degrees of freedom. By combining YOLO11n-OBB with calibrated pixel-to-robot geometry, the system chooses between two simple motion policies. It uses a direct grasp for utensils within 45 degrees of the radial line from the robot origin to the utensil center, and a sweep-then-grasp strategy for utensils beyond that angle. Trials were run one utensil at a time on a controlled tabletop workspace, with each of the seven utensils placed across nine 10-degree starting-angle ranges from 0 to 90 degrees. Across 326 logged motions, 196 were direct grasps, and 130 were corrective sweep passes, while a lightweight HSV/LAB heuristic labeled color correctly in at least 91\% of detections for the four in-gamut utensils. In total, 176 of the 196 trials (89.8\%) ended with the utensil in the bin for its color. While the final performance still depends on the detector's quality and the gripper's mechanics, in this task and setup the corrective sweeps brought every steeply misaligned utensil into a graspable orientation without an added wrist joint, keeping the lower cost and simplicity of 4-DoF hardware. These results are limited to the seven tested utensils, which also appeared in the detector's training images, and to a single camera and a painted wooden tabletop, with manually triggered trials and no direct-grasp-only baseline. The accuracy of the detected orientation and of the pixel-to-robot mapping was checked only against the system's own images and logs and not against an external reference. The next validation step is the one named under Future directions, the same angle protocol on held-out utensils in randomized order, with and without corrective sweeps, and the orientation and pixel-to-robot errors measured directly.

\hypertarget{conflict-of-interest}{%
\section{\texorpdfstring{\textbf{CONFLICT OF
INTEREST}}{CONFLICT OF INTEREST}}\label{conflict-of-interest}}

The authors declare no conflicts of interest related to this work.

\hypertarget{ai-disclosure-statement}{%
\section{AI DISCLOSURE STATEMENT}\label{ai-disclosure-statement}}

During the preparation of this work, the author(s) used Claude
(Anthropic) for programming assistance, Grammarly for language editing,
and Canva for image editing (figure labeling and noise removal).
Following the use of these tools, the author(s) carefully reviewed,
verified, and edited the content as necessary and take full
responsibility for the accuracy, originality, and integrity of the
published work.

\hypertarget{references}{%
\section{\texorpdfstring{\textbf{REFERENCES}}{REFERENCES}}\label{references}}

1. Waveshare. RoArm-M2-S. Available from: \url{https://www.waveshare.com/roarm-m2-s.htm} (accessed on 2025-08-12)

2. Waveshare. RoArm-M2-S wiki. Available from: \url{https://www.waveshare.com/wiki/RoArm-M2-S} (accessed on 2025-08-12)

3. Lynch KM, Park FC. Modern robotics: mechanics, planning, and control. Cambridge: Cambridge University Press; 2017. ISBN 978-1-107-15630-2.

4. Mason MT. Mechanics and planning of manipulator pushing operations. Int J Robot Res. 1986;5(3):53-71. doi:10.1177/027836498600500303.

5. Chang LY, Srinivasa SS, Pollard NS. Planning pre-grasp manipulation for transport tasks. In: Proceedings of the IEEE International Conference on Robotics and Automation (ICRA); 2010. p. 2697-2704. doi:10.1109/ROBOT.2010.5509651.

6. Dogar MR, Srinivasa SS. Push-grasping with dexterous hands: mechanics and a method. In: Proceedings of the IEEE/RSJ International Conference on Intelligent Robots and Systems (IROS); 2010. p. 2123-2130. doi:10.1109/IROS.2010.5652970.

7. Zeng A, Song S, Welker S, Lee J, Rodriguez A, Funkhouser T. Learning synergies between pushing and grasping with self-supervised deep reinforcement learning. In: Proceedings of the IEEE/RSJ International Conference on Intelligent Robots and Systems (IROS); 2018. p. 4238-4245. doi:10.1109/IROS.2018.8593986.

8. Hang K, Morgan AS, Dollar AM. Pre-grasp sliding manipulation of thin objects using soft, compliant, or underactuated hands. IEEE Robot Autom Lett. 2019;4(2):662-669. doi:10.1109/LRA.2019.2892591.

9. Chaumette F, Hutchinson S. Visual servo control. I. Basic approaches. IEEE Robot Autom Mag. 2006;13(4):82-90. doi:10.1109/MRA.2006.250573.

10. Cong VD. Visual servoing control of 4-DOF palletizing robotic arm for vision based sorting robot system. Int J Interact Des Manuf. 2023;17(2):717-728. doi:10.1007/s12008-022-01077-8.

11. Rangarajan A. 4DoF vision robotic pen sorting. GitHub. Available from: \url{https://github.com/Anirudhpro/4DoF_vision_robotic_pen_sorting} (commit d76d720; accessed on 2026-10-07)

12. Zhang Z. A flexible new technique for camera calibration. IEEE Trans Pattern Anal Mach Intell. 2000;22(11):1330-1334.

13. OpenCV. Detection of ArUco markers. OpenCV Documentation, version 4.x. Available from: \url{https://docs.opencv.org/4.x/d5/dae/tutorial_aruco_detection.html} (accessed on 2025-08-12)

14. Garrido-Jurado S, Muñoz-Salinas R, Madrid-Cuevas FJ, Marín-Jiménez MJ. Automatic generation and detection of highly reliable fiducial markers under occlusion. Pattern Recognit. 2014;47(6):2280-2292. doi:10.1016/j.patcog.2014.01.005.

15. Jocher G, Qiu J. Ultralytics YOLO11. Available from: \url{https://github.com/ultralytics/ultralytics} (accessed on 2026-06-08)

16. Rangarajan A. pen-hp5la [Dataset]. Roboflow Universe. Available from: \url{https://universe.roboflow.com/tbn-acjk2/pen-hp5la/dataset/1} (accessed on 2026-10-07)

17. Google LLC. Google Colaboratory. Available from: \url{https://colab.research.google.com} (accessed on 2026-06-08)

18. Ultralytics. Oriented bounding boxes (OBB). Ultralytics Docs. Available from: \url{https://docs.ultralytics.com/tasks/obb/} (accessed on 2026-06-08)

19. Rangarajan A. Pens, version 2 [Dataset]. Roboflow Universe. Available from: \url{https://universe.roboflow.com/tbn-acjk2/pens-y8dxo/dataset/2} (accessed on 2026-10-07)

20. egogor. pen [Dataset]. Roboflow Universe. Available from: \url{https://universe.roboflow.com/egogor/pen-jw4m2} (accessed on 2026-10-07)

\end{document}